\documentclass[conference]{IEEEtran}
\IEEEoverridecommandlockouts
\usepackage{cite}
\usepackage{amsmath,amssymb,amsfonts}
\usepackage{algorithmic}
\usepackage{graphicx}
\usepackage{textcomp}
\usepackage{xcolor}
\def\BibTeX{{\rm B\kern-.05em{\sc i\kern-.025em b}\kern-.08em
    T\kern-.1667em\lower.7ex\hbox{E}\kern-.125emX}}
\begin{document}

\title{PeP: a Point enhanced Painting method for unified point cloud tasks \\
{\footnotesize \textsuperscript{*}Note: Technical Report Version}
}

\author{\IEEEauthorblockN{1\textsuperscript{st} Zichao Dong}
\IEEEauthorblockA{\textit{UDeer.ai} \\
zichao@udeer.ai}
\and
\IEEEauthorblockN{2\textsuperscript{nd} Hang Ji}
\IEEEauthorblockA{\textit{UDeer.ai} \\
jihang@udeer.ai
}
\and
\IEEEauthorblockN{3\textsuperscript{rd} Xufeng Huang}
\IEEEauthorblockA{\textit{UDeer.ai} \\
xufeng@udeer.ai
}
\and
\IEEEauthorblockN{4\textsuperscript{th} Weikun Zhang}
\IEEEauthorblockA{\textit{Zhejiang University} \\
zhangwk@zju.edu.cn
}
\and
\IEEEauthorblockN{5\textsuperscript{th} Xin Zhan}
\IEEEauthorblockA{\textit{UDeer.ai} \\
zhanxin@udeer.ai}
\and
\IEEEauthorblockN{6\textsuperscript{th} Junbo Chen *}
\IEEEauthorblockA{\textit{UDeer.ai} \\
junbo@udeer.ai}

\thanks{This is a draft version contains nuScenes lidar segmentation benchmark only.}

}

\maketitle

\begin{abstract}  

Point encoder is of vital importance for point cloud recognition. As the very beginning step of whole model pipeline, adding features from diverse sources and providing stronger feature encoding mechanism would provide better input for downstream modules. In our work, we proposed a novel PeP module to tackle above issue. PeP contains two main parts, a refined point painting method and a LM-based point encoder. Experiments results on the nuScenes and KITTI datasets validate the superior performance of our PeP. The advantages leads to strong performance on both semantic segmentation and object detection, in both lidar and multi-modal settings. Notably, our PeP module is model agnostic and plug-and-play. Our code will be publicly available soon.

\end{abstract}


\section{Introduction}
Point cloud recognition is an important yet challenging perception task, which is widely used in industry especially in autonomous driving area. Previous feature enhancement module mainly rely on semantic segmentation on paired image associated with point cloud frames to operate point painting step. However, with the rapid development of computer vision, open vocabulary segmentation method like ODISE\cite{xu2023open} would provide stunning panoptic segmentation result. With the help of stronger segmentors, instance id and detailed semantic label would be used as a semantic prior for lidar point cloud perception. We also notice that even without paired image, lidar semantic segmentation models could paint themselves in a two-stage pipeline, which would also boost their accuracy.  On the other hand, there are rich yet growing literature investigating in combine LM model with point cloud perception. Nevertheless, rare work attempts to define point cloud recognition as a language modeling issue. In our PeP, each point is considered as a sentence, who is constructed by several word like attributes(like lidar raw input value). Inspired by word embedding mechanism, a tokenizer is also utilized to embed attributes features from scalar to a learned embedding. Similar to NLP, self-attention module is used to let attribute embeddings communicate with each other within a single point. Delightedly, our LM-based point encoding layer would boost model when seperately used, while get larger gain accompany with our refined point painting module. 

Our contributions can be summarized as follows:  

1. A more flexible and accurate point painting method is designed, which is validated in both detection and semenatic segmentation task. 

2. A LM-based point encoder is utilized to extract strong embedding for points. 

3. Without bells and whistles, our PeP achives state-of-the-art performance on both lidar semantic segmentation task and multi-modal 3d object detection task.

\section{RELATED WORK}
\subsection{Point cloud segmentation}
Point cloud segmentation is the process of classifying points into distinct and homogeneous groups \cite{xiao2023position}. In the context of outdoor scenes, point clouds generated by LiDAR technology often exhibit sparsity and non-uniform distribution. This specific segmentation challenge is commonly referred to as LiDAR-based 3D segmentation. LiDAR-Based 3D Semantic Segmentation involves the classification of point clouds based on their semantic characteristics. Depending on how point clouds are represented, existing approaches can be categorized into three main streams \cite{xiao2023position}: point-based methods \cite{qi2017pointnet,qi2017pointnet++}, voxel/grid-based methods, and multi-modal methods \cite{xu2021rpvnet,tang2020searching}. Point-based methods focus on processing individual points, while voxel/grid-based methods discretize point clouds into 3D voxels or 2D grids and apply convolutional operations. Our approach, PeP, falls under the multi-modal method category. In our work, we introduce a novel point enhanced painting method designed to address various tasks related to point clouds in a unified manner.

\subsection{Strong baseline: SphereFormer}
Lai \textit{et al.} \cite{lai2023spherical} delve into the distribution of varying sparsity among LiDAR points. Their approach, SphereFormer, aims to directly gather information from densely packed points to the sparsely scattered ones \cite{lai2023spherical}. Lai \textit{et al.} introduced a novel concept known as radial window self-attention, which effectively divides the space into distinct, non-overlapping, elongated windows. This method successfully mitigates connectivity issues while substantially expanding the scope of information gathering, resulting in a remarkable enhancement in the performance of sparsely distributed points that are far from the sensor \cite{lai2023spherical}.Furthermore, to adapt to these elongated windows, Lai \textit{et al.} introduced the concept of exponential splitting for generating finely detailed position encoding. They also employ dynamic feature selection to augment the model's representational capacity \cite{lai2023spherical}.
% Xiao \textit{et al.} \cite{xiao2023position} proposed P3Former, which uses a specialized positional embedding to guide the whole segmentation procedure \cite{xiao2023position}. The authors initially devised a Hybrid Parameterized Positional Embedding, which combines both polar and Cartesian spatial representations. MPE merges the polar coordinate pattern prior to point distribution with Cartesian embeddings, resulting in a robust embedding that forms the foundation for position-guided segmentation. It is first incorporated into backbone features to serve as a positional discriminator, effectively distinguishing instances with similar geometric characteristics. Additionally, it is applied to mask prediction and masked cross-attention, leading to Position-Aware Segmentation and Masked Focal Attention. PA-Seg introduces a parallel branch for position-based mask prediction alongside the original feature-based method, compensating for the absence of absolute positional information in high-level features. MFA simplifies the masked attention operation by replacing the cross-attention map with the integrated mask prediction. These design choices collectively enable queries to focus on specific positions and predict small masks within a defined region \cite{xiao2023position}.

\subsection{Point cloud detection}
The goal of 3D object detection is to predict rotated bounding boxes in three dimensions \cite{yin2021center}. In contrast to 2D detectors, 3D detectors differ in their input encoding approach. Vote3Deep \cite{engelcke2017vote3deep} employs feature-centric voting to efficiently process sparse 3D point clouds within equally spaced 3D voxels. VoxelNet \cite{zhou2018voxelnet} utilizes a PointNet \cite{qi2017pointnet} within each voxel to generate a unified feature representation. This representation is then used in a head that employs both 3D sparse convolutions and 2D convolutions to produce detections \cite{yin2021center}. PIXOR \cite{yang2018pixor} projects all points onto a 2D feature map, incorporating 3D occupancy and point intensity data to eliminate the computational cost of 3D convolutions. PointPillars \cite{lang2019pointpillars} replaces voxel computations with a pillar representation, essentially a single tall elongated voxel at each map location, enhancing the efficiency of the backbone.

\subsection{Strong baseline: centerpoint}
Yin \textit{et al.} \cite{yin2021center} introduced a two-stage object detection method called CenterPoint. In the two-stage 3D detector CenterPoint, the first stage utilizes a keypoint detector to locate the centers of objects and their attributes \cite{yin2021center}. Specifically, CenterPoint employs a standard LiDAR-based backbone network to construct a representation of the input point cloud \cite{yin2021center}. This representation is then flattened into an overhead map view, and a conventional image-based key point detector is employed to identify object centers. For each detected center, it regresses to determine other object properties, such as three-dimensional dimensions, orientation, and velocity, based on point features at the center location. Additionally, a lightweight second stage is used to enhance object positions, which extracts point features at the three-dimensional centers of each face of the estimated objects' 3D bounding boxes. This second stage helps recover local geometric information that may be lost due to striding and a limited receptive field, resulting in a substantial performance improvement at a relatively low computational cost. This approach offers several advantages, including reduced search space due to the absence of intrinsic orientation in point-based representations, simplification of downstream tasks like tracking, and efficient two-stage refinement modules for faster processing \cite{yin2021center}.

\subsection{Multi-modality 3d object detection}
Multi-modal 3D Object Detection integrates data from both Lidar and color cameras \cite{yin2021multimodal}. In the case of Frustum PointNet \cite{qi2017pointnet} and Frustum ConvNet \cite{wang2019frustum}, the initial step involves object detection in the image space to identify regions of interest within the point cloud for subsequent processing. This approach enhances the efficiency and precision of 3D detection ; however, its effectiveness is inherently reliant on the quality of 2D detections \cite{yin2021multimodal}. MVX-Net \cite{sindagi2019mvx}, PointAugmenting \cite{wang2021pointaugmenting}, and PointPainting \cite{vora2020pointpainting} leverage point-wise correspondence to annotate each Lidar point with image-based segmentation or CNN features. 

\subsection{Strong baseline: Virconv}
Wu \textit{et al.} \cite{wu2023virtual} have proposed a VirCon vNet pipeline based on a novel Virtual Sparse Convolution operator \cite{wu2023virtual}. The design of this approach is grounded in two key observations \cite{wu2023virtual}. Firstly, in LiDAR scans, the geometries of nearby objects tend to be relatively complete. Consequently, most virtual points from nearby objects provide only marginal performance improvement, but significantly increase computational costs \cite{wu2023virtual}. Secondly, noisy points resulting from inaccurate depth completions are predominantly distributed along instance boundaries. Once projected onto the image plane, these noisy points can be identified in 2D images \cite{wu2023virtual}. Building on these insights, the authors have devised a Stochastic Voxel Discard scheme, which preserves the most crucial virtual points through bin-based sampling. This involves discarding a substantial number of nearby voxels while retaining distant ones, leading to a substantial acceleration in network computation \cite{wu2023virtual}. Additionally, they have introduced a Noise-Resistant Submanifold Convolution layer to encode geometric features of voxels in both 3D space and 2D image space. The extended receptive field in 2D space enables NRConv to differentiate noise patterns along instance boundaries in 2D image space. As a result, the detrimental effects of noise can be mitigated \cite{wu2023virtual}.

\subsection{Previous point painting method}
% PointPainting \cite{vora2020pointpainting} represents a straightforward yet highly efficient method for sequential fusion. This technique involves projecting each lidar data point onto the output generated by an image semantic segmentation network \cite{vora2020pointpainting}. Following this projection, the channel-wise activations are seamlessly incorporated into the intensity measurements of each respective lidar point. These amalgamated (or "painted") lidar points can subsequently be seamlessly integrated into any lidar detection approach. PointPainting effectively mitigates the limitations inherent in previous fusion concepts \cite{vora2020pointpainting}. Unlike its predecessors, PointPainting imposes no constraints on the underlying 3D detection architecture \cite{vora2020pointpainting}. Moreover, it effectively eliminates the issues associated with feature and depth blurring. This technique doesn't necessitate the computation of a pseudo-point cloud, and it places no restrictions on the maximum achievable recall rate \cite{vora2020pointpainting}.

MVP \cite{yin2021multimodal} is a straightforward yet effective framework designed for the fusion of 3D LiDAR and high-resolution color measurements. Yin \textit{et al.} \cite{yin2021multimodal} utilizes close-range depth measurements from the LiDAR sensor to project RGB measurements into the scene, thereby enhancing the RGB measurements into three-dimensional virtual points \cite{yin2021multimodal}. The proposed multi-modal virtual point detector, MVP, is capable of generating high-resolution three-dimensional point clouds near target objects. Subsequently, a center-based 3D detector identifies all objects within the scene. To elaborate, MVP employs 2D object detection to segment the original point cloud into instance frustums. Following this, MVP generates densely-packed three-dimensional virtual points near these foreground points by transforming two-dimensional pixels into three-dimensional space. Depth completion in the image space is utilized to infer the depth of each virtual point. Ultimately, MVP combines these virtual points with the original LiDAR measurements as input for a conventional center-based 3D detector \cite{yin2021multimodal}.

\subsection{ODISE}
ODISE \cite{xu2023open} is a Diffusion-based Open-vocabulary DIffusion-based panoptic SEgmentation model, which employs both large-scale text-image diffusion and discriminative models to achieve cutting-edge panoptic segmentation for any category in various real-world scenarios \cite{xu2023open}. The model's approach consists of several components. Initially, it includes a pre-trained and fixed text-to-image diffusion model. In this model, images along with their captions are input, and internal features of the diffusion model are extracted for them. Utilizing these features as input, ODISE's mask generator can produce panoramic masks encompassing all conceivable concepts within the image. The mask generator is trained using annotated masks available in the training dataset. Subsequently, a mask classification module categorizes each mask into one of numerous open-vocabulary categories by associating the diffusion features of each predicted mask with text embeddings of several object category names \cite{xu2023open}. This classification module is trained with either mask category labels or image-level captions from the training dataset. Once the training is complete, Xu \textit{et al.} \cite{xu2023open} employ both the text-image diffusion model and the discriminative model to execute open-vocabulary panoptic inference, classifying predicted masks.

\subsection{Transformer based LM}
BERT \cite{devlin2018bert} is a bidirectional pre-trained language model that predicts words in the input text by using masked language modeling. BERT has performed exceptionally well in various natural language processing tasks and has become the foundation for many NLP applications. GPT \cite{floridi2020gpt} is a transformer-based autoregressive language model that generates text in a left-to-right manner. It is commonly used for text generation, conversation, and similar tasks. Llama \cite{touvron2023llama} is also based on the Transformer architecture. The distinctive feature of this model is its training on a larger number of tokens, resulting in a series of language models that perform optimally across various inference budgets. The Llama model, with 130 billion parameters, outperforms GPT-3 on most benchmarks and can run on a single V100 GPU.

\subsection{PointNet}
PointNet \cite{qi2017pointnet} is a comprehensive architectural framework designed to take point clouds as direct input, generating outputs that can encompass either class labels for the entire input or segment/part labels for each individual point within the input data \cite{qi2017pointnet}. In the basic configuration, each point is solely represented by its three-dimensional coordinates \cite{qi2017pointnet}. The potential for additional dimensions arises from the computation of normals and the inclusion of other local or global features. A pivotal aspect of our methodology involves the strategic application of max pooling \cite{qi2017pointnet}. In essence, the network acquires proficiency in learning a set of optimization functions and criteria, which, in turn, facilitate the selection of intriguing or informative points from the point cloud, and encode the rationale behind such selections \cite{qi2017pointnet}. The final layers of full connectivity in the network consolidate these learned optimal values into a global descriptor for the entire shape or employ them to predict labels for each individual point \cite{qi2017pointnet}.

\section{METHOD}
\subsection{Overview}
\begin{figure*}[htbp]
    \centering
    \includegraphics[width=16cm]{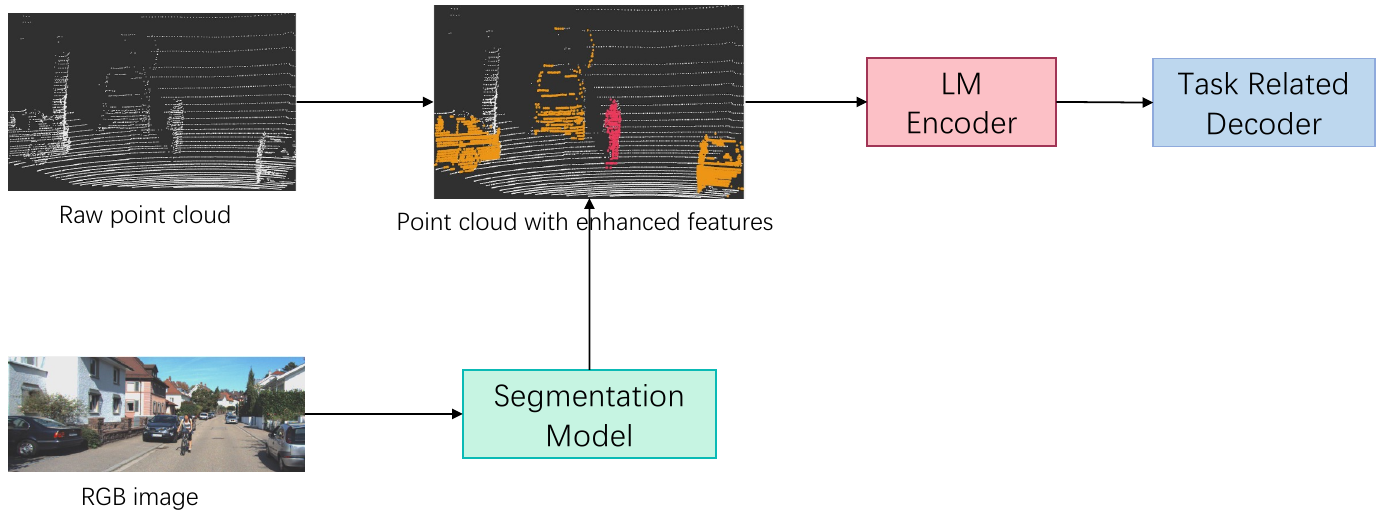}
    \caption{\small \textbf{Overall pipeline of PeP}. Our model takes both the image and point cloud as inputs. The image is processed by the segmentation model to extract semantic information. Subsequently, this semantic information is applied to the point cloud through painting, enhancing its features. The enriched point cloud is then forwarded to the language model (LM), followed by a task related decoder for further processing.}
    \label{fig:model}
\end{figure*}

Our PeP is mainly constructed by two main components, LM-based point encoder and point painting respectively. The above two components cloud boost model by single while combining them together would harvest more gain. Below section would dipict them in detail. The pipline of PeP is shown in Fig.~\ref{fig:model}.

\subsection{LM-based point encoder}
As we all know, language model has shown his strong ability in encoding sequence inputs. Inspired by LM, we define traditional point cloud encoding problem as a sequence encoding problem just like NLP. To be specific, a point cloud could be consider as a paragraph who is constructed by a set of independent points just like sentences. Further, a sentence-like point cloud also be considered as a combination of word-like attributes, such as x,y,z values. As for language models, a word encoder would be used to encode word as a word embedding while words would operate self attention in transformer based encoder.   

In our LM-based point encoder, a attribute encoding layer would first take each single one original input feature(like x,y,z) as an input and embedding it as a n-dim feature. After this step, the original m-dim input feature would be encoded a n*m dimension feature. Though there is trivial computation cost, we find it could benefit a lot for various point cloud recognition tasks.  

As for the reasons, we think stronger feature encoding capacity and modality alignment would be potential causes. Firstly, compared with a single scalar, a learned n-dim embedding would naturally have stronger feature expression ability. Take LM as an example, word embedding layer plays an important while naive one-hot tokenizer is just like original scalar feature in lidar points. In our experiments, different from word embedding in LM, we find that our embedding dimension could as small as 4, while large embedding dimension would make whole model hard to converge. On the other hand, different modality and various dimension cloud be found in point features. For instance, x indicates distance with meter as measurement while t stands for timestamps with second as measurement. We also notice that some previous works also do some hand-craft normalization trick with magic numbers(like dived 10 for intensity value). The normalization step would be done in a learning based way in our LM-based point encoder. Apart from that, we hope that PeP could also served as a modality aligner to unify features from separate modality. Thanks to self attention layer, aligned features from separate attributes would have deeper communications. It is worth mentioning that this module could be used in various tasks while ignoring downstream module as a point feature encoder. 

Detailed experimental result would be shown in below quantitative evaluation section for tasks including lidar object detection, multi-modality object detection and lidar semantic segmentation. 

\subsection{Point painting for segmentation tasks}
As for original point painting, a vision-based segmenter is used to assign semantic label to each points in point cloud. However, we find it is straight forward to paint points by lidar segmentation model itself. As far as we are concerned, there are two main advantages as below.  

Firstly, our model would serve as a two stage self-correction segmentor by our design. For the first stage, no semantic label is known, thus all points are painted with semantic label -1. In this stage, our model would segment point clouds with no prior as previous works. After stage one, we get a coarse semantic label for each points. By painting orignal points with results in stage one, our model would inference again with semantic prior. Similar to COT, our model would self-correct itself with previous inference results. 

Secondly, there would be no confusing noise for point cloud caused by timestamp and calibration error. Take original point painting as an example, each lidar point would be projected in image coordination while be assinged same semantic label as the corresponding projected pixel. However, the semantic label may not be extreme accurate while there would also be projection error. Thus, 3d lidar point would be mis-painted, which is hard for downstream model to denoise since neighbours in projected 2d space may far away in 3d space. 

It is worth mentioning that our method is model agnostic, which could also be applied to any existing lidar segmentation model. 

\subsection{Point painting for multi-modality 3d object detection}
As for multi-modality 3d object detection task, we use a diffusion based open vocabulary instance segmentation model to provide semantic prior. For each pixel, we could acquire semantic label and a unique instance id as clue for lidar point painting step. Different from original version point painting which only have same labels as final classes, our open vocabulary based segmentor cloud provide accurate sub-class information and attach a label for background pixels. In our experiments, we find refined classes and dense prediction panoptic segmentation label could provide more vision prior compared with foreground segmentation. In automatic driving scenes, occlusion is common which may be a burden for object detection task. To tackle above issue, we notice that a instance id for each point is an effective prior.  

However, as mentioned in previous section, inaccurate segmentation result and noisy projection between lidar and image would confuse 3d lidar point cloud perception. To alleviate this issue, we use VirConv as our baseline, whose NRConv provides a valid method to denoise 3d points considering 2d relative position. To be precise, compared with RGB value painted on points in VirConv, semantic label and instance id are also appended. We believe that recent work like UniTr who has similar denoising mechanism cloud also be a good choice.

\subsection{Combine point painting with LM-based point encoder}
Separate LM-based point encoder and point painting would both boost our perception model, while combining them together would have larger gain. Semantic label and instance id are high level features while xyz value are raw sensor outputs. Our point encoder would align them and provide a point-wise way to fuse attributes from different sources. We believe that point painting is a simple way to add a specific feature to a point, while more reliable features would naturally be beneficial. On the other hand, LM-based point encoder could provide stronger encoding mechanism ignoring specific input. The two module are orthogonal in design while cascade in usage. 

\section{Experiment}

\subsection{Dataset}
We validate the proposed PeP on both KITTI dataset and nuScenes dataset.
\subsubsection{KITTI object detection}
The KITTI 3D object detection dataset \cite{geiger2012we} comprises a total of 7,481 LiDAR and image frames for training and 7,518 for testing purposes. The training data has been split into 3,712 frames for training and 3,769 frames for validation, following the approach of recent studies. We also employed the widely accepted evaluation metric: 3D Average Precision (AP) with 40 recall thresholds (R40). Specifically, the IoU thresholds for cars, pedestrians, and cyclists are set at 0.7, 0.5, and 0.5, respectively. It's worth noting that we used the KITTI odometry dataset as a substantial unlabeled dataset, which consists of 43,552 LiDAR and image frames. From this dataset, we uniformly sampled 10,888 frames, referred to as the semi dataset, and utilized them for training our VirConv-S model. Importantly, after cross-referencing the mapping files, we found no overlapping data between the KITTI 3D detection dataset and the KITTI odometry dataset.

\subsubsection{nuScenes lidar semantic segmentation}
nuScenes \cite{caesar2020nuscenes} stands as a well-known multimodal dataset tailored for 3D object detection within urban environments. Comprising 1000 distinct driving sequences, each spanning 20 seconds, the dataset boasts meticulous 3D bounding box annotations. The Lidar operates at a frequency of 20Hz, and while it offers sensor and vehicle pose data for each Lidar frame, object annotations occur only every ten frames, equating to a 0.5-second interval. The dataset goes to great lengths to safeguard privacy, obscuring any personally identifiable information and pixelating faces and license plates in the color images. Impressively, it incorporates a total of six RGB cameras, boasting a resolution of 1600 × 900 and capturing data at a rate of 12Hz. As for Lidar semgemntation task, we follow panoptic nuScenes \cite{fong2021panoptic} setting.

\subsection{Implementation Details}
\subsubsection{Point painting for lidar semantic segmentation} Only semenatic label for each points is added with original features. Assuming that original points feature like [feats], for first time inference points would be [feats, -1] while second time like [feats, lbl]. The lbl denotes results for each points in stage one. By the way, the range of lbl stays unchanged with segmentation task. 

\subsubsection{Point painting for multi-modal 3d object detection} A open-vocabulary segmentation model is used to produce semantic and instance label for each points. For KITTI object detection task, our class label prompt including [person; car; bicycle; road; building; vegetation; wall; tree; plant; sidewalk; grass; truck; van; motorcycle; house; stone; fence; traffic light; rail; pole; train; traffic sign]. Assuming that original points feature like [feats], painted points like [feats, semantic-lbl, instance-id]. 

\subsubsection{Settings for LM-based point encoder} Assuming a point cloud has n points and m features for each points, the input shape should be [n, m]. Firstly, a tokenizer would encode it to [n, m, d]. We find d=4 is a satisfying parameter. Subsequently, a single layer self-attention layer is used as an encoder.

\subsection{Quantitive Evaluation}
The quantitive results on KITTI dataset are shown in Table \ref{table:table1}. The quantitive results on NuScenes dataset are shown in Table \ref{table:table2}. 

\begin{table}[htbp]
\centering
\caption{3D AP(R40) of Car on the val set of KITTI dataset.}
\label{table:table1}
\begin{tabular}{llll}
\hline
                           & \textbf{Easy} & \textbf{Medium} & \textbf{Hard} \\ \hline
\textbf{Baseline(VirConv)} & 95.76         & 90.91           & 88.61         \\
\textbf{Ours}              & 95.57         & 90.98           & 91.01         \\ \hline
\end{tabular}
\end{table}

\begin{table*}[htbp]
\centering
\caption{The results on NuScenes validation set with TTA.}
\label{table:table2}
\vspace{4mm}
% \footnotesize
\centering
\tabcolsep 3pt
\begin{tabular}{lccccccccccccccccc}
\hline
\textbf{}                       & \textbf{mIOU} & \rotatebox{90}{\textbf{barrier}} & \rotatebox{90}{\textbf{bicycle}} & \rotatebox{90}{\textbf{bus}} & \rotatebox{90}{\textbf{car}} & \rotatebox{90}{\textbf{construction}} & \rotatebox{90}{\textbf{motocycle}} & \rotatebox{90}{\textbf{pedstrain}} & \rotatebox{90}{\textbf{Traffic cone}} & \rotatebox{90}{\textbf{trailer}} & \rotatebox{90}{\textbf{truck}} & \rotatebox{90}{\textbf{drivable}} & \rotatebox{90}{\textbf{Other flat}} & \rotatebox{90}{\textbf{sidewalk}} & \rotatebox{90}{\textbf{terrian}} & \rotatebox{90}{\textbf{manmade}} & \rotatebox{90}{\textbf{vegetation}} \\ \hline
\textbf{Baseline(SphereFormer)} & 78.4          & 77.7             & 43.8             & 94.5         & 93.1         & 52.4                  & 86.9               & 81.2               & 65.4                  & 73.4             & 85.3           & 97                & 73.4                & 75.4              & 75               & 91               & 89.2                \\
\textbf{Ours}                   & 79.8          & 78.6             & 47               & 95.1         & 93.9         & 53.9                  & 89                 & 81.3               & 67.9                  & 73.9             & 87             & 97.1              & 74.1                & 76,2              & 75.7             & 91               & 90.2                \\ \hline
\end{tabular}
\end{table*}

\subsection{Discussion}
In our experiments, we find that the more clue we provide the high accuracy we could achieve. Thus, we consider our PeP is a expandable framework for feature potential useful clues. However, we also find it is relatively difficult to converge for KITTI, which indicates that LM-based point encoder may be hungry for large amount of data. Further, with help of our PeP, point cloud perception pipeline would be data dependant at very beginning.

\section*{Future work}
Our PeP is just a simple attempt in combining language model with point cloud recognition. In the future, we would try to combine modern language model to point cloud. Apart from that, we notice that recent diffusion based method could extract distinctive features from
 images. Later we may try to paint points with image features rather that result level semantic and instance results. Last but certainly not the least, since point cloud classification task is essential in point cloud recognition, we may try to marriage our LM-based point encoder with point cloud classification in our future work.

\bibliography{references}

\begin{thebibliography}{10}

\bibitem{caesar2020nuscenes}
Holger Caesar, Varun Bankiti, Alex~H Lang, Sourabh Vora, Venice~Erin Liong, Qiang Xu, Anush Krishnan, Yu~Pan, Giancarlo Baldan, and Oscar Beijbom.
\newblock nuscenes: A multimodal dataset for autonomous driving.
\newblock In {\em Proceedings of the IEEE/CVF conference on computer vision and pattern recognition}, pages 11621--11631, 2020.

\bibitem{devlin2018bert}
Jacob Devlin, Ming-Wei Chang, Kenton Lee, and Kristina Toutanova.
\newblock Bert: Pre-training of deep bidirectional transformers for language understanding.
\newblock {\em arXiv preprint arXiv:1810.04805}, 2018.

\bibitem{engelcke2017vote3deep}
Martin Engelcke, Dushyant Rao, Dominic~Zeng Wang, Chi~Hay Tong, and Ingmar Posner.
\newblock Vote3deep: Fast object detection in 3d point clouds using efficient convolutional neural networks.
\newblock In {\em 2017 IEEE International Conference on Robotics and Automation (ICRA)}, pages 1355--1361. IEEE, 2017.

\bibitem{floridi2020gpt}
Luciano Floridi and Massimo Chiriatti.
\newblock Gpt-3: Its nature, scope, limits, and consequences.
\newblock {\em Minds and Machines}, 30:681--694, 2020.

\bibitem{fong2021panoptic}
Whye~Kit Fong, Rohit Mohan, Juana~Valeria Hurtado, Lubing Zhou, Holger Caesar, Oscar Beijbom, and Abhinav Valada.
\newblock Panoptic nuscenes: A large-scale benchmark for lidar panoptic segmentation and tracking.
\newblock {\em arXiv preprint arXiv:2109.03805}, 2021.

\bibitem{geiger2012we}
Andreas Geiger, Philip Lenz, and Raquel Urtasun.
\newblock Are we ready for autonomous driving? the kitti vision benchmark suite.
\newblock In {\em 2012 IEEE conference on computer vision and pattern recognition}, pages 3354--3361. IEEE, 2012.

\bibitem{lai2023spherical}
Xin Lai, Yukang Chen, Fanbin Lu, Jianhui Liu, and Jiaya Jia.
\newblock Spherical transformer for lidar-based 3d recognition.
\newblock In {\em CVPR}, 2023.

\bibitem{lang2019pointpillars}
Alex~H Lang, Sourabh Vora, Holger Caesar, Lubing Zhou, Jiong Yang, and Oscar Beijbom.
\newblock Pointpillars: Fast encoders for object detection from point clouds.
\newblock In {\em Proceedings of the IEEE/CVF conference on computer vision and pattern recognition}, pages 12697--12705, 2019.

\bibitem{qi2017pointnet}
Charles~R Qi, Hao Su, Kaichun Mo, and Leonidas~J Guibas.
\newblock Pointnet: Deep learning on point sets for 3d classification and segmentation.
\newblock In {\em Proceedings of the IEEE conference on computer vision and pattern recognition}, pages 652--660, 2017.

\bibitem{qi2017pointnet++}
Charles~Ruizhongtai Qi, Li~Yi, Hao Su, and Leonidas~J Guibas.
\newblock Pointnet++: Deep hierarchical feature learning on point sets in a metric space.
\newblock {\em Advances in neural information processing systems}, 30, 2017.

\bibitem{sindagi2019mvx}
Vishwanath~A Sindagi, Yin Zhou, and Oncel Tuzel.
\newblock Mvx-net: Multimodal voxelnet for 3d object detection.
\newblock In {\em 2019 International Conference on Robotics and Automation (ICRA)}, pages 7276--7282. IEEE, 2019.

\bibitem{tang2020searching}
Haotian Tang, Zhijian Liu, Shengyu Zhao, Yujun Lin, Ji~Lin, Hanrui Wang, and Song Han.
\newblock Searching efficient 3d architectures with sparse point-voxel convolution.
\newblock In {\em European conference on computer vision}, pages 685--702. Springer, 2020.

\bibitem{touvron2023llama}
Hugo Touvron, Thibaut Lavril, Gautier Izacard, Xavier Martinet, Marie-Anne Lachaux, Timoth{\'e}e Lacroix, Baptiste Rozi{\`e}re, Naman Goyal, Eric Hambro, Faisal Azhar, et~al.
\newblock Llama: Open and efficient foundation language models.
\newblock {\em arXiv preprint arXiv:2302.13971}, 2023.

\bibitem{vora2020pointpainting}
Sourabh Vora, Alex~H Lang, Bassam Helou, and Oscar Beijbom.
\newblock Pointpainting: Sequential fusion for 3d object detection.
\newblock In {\em Proceedings of the IEEE/CVF conference on computer vision and pattern recognition}, pages 4604--4612, 2020.

\bibitem{wang2021pointaugmenting}
Chunwei Wang, Chao Ma, Ming Zhu, and Xiaokang Yang.
\newblock Pointaugmenting: Cross-modal augmentation for 3d object detection.
\newblock In {\em Proceedings of the IEEE/CVF Conference on Computer Vision and Pattern Recognition}, pages 11794--11803, 2021.

\bibitem{wang2019frustum}
Zhixin Wang and Kui Jia.
\newblock Frustum convnet: Sliding frustums to aggregate local point-wise features for amodal 3d object detection.
\newblock In {\em 2019 IEEE/RSJ International Conference on Intelligent Robots and Systems (IROS)}, pages 1742--1749. IEEE, 2019.

\bibitem{wu2023virtual}
Hai Wu, Chenglu Wen, Shaoshuai Shi, Xin Li, and Cheng Wang.
\newblock Virtual sparse convolution for multimodal 3d object detection.
\newblock In {\em Proceedings of the IEEE/CVF Conference on Computer Vision and Pattern Recognition}, pages 21653--21662, 2023.

\bibitem{xiao2023position}
Zeqi Xiao, Wenwei Zhang, Tai Wang, Chen~Change Loy, Dahua Lin, and Jiangmiao Pang.
\newblock Position-guided point cloud panoptic segmentation transformer.
\newblock {\em arXiv preprint arXiv:2303.13509}, 2023.

\bibitem{xu2021rpvnet}
Jianyun Xu, Ruixiang Zhang, Jian Dou, Yushi Zhu, Jie Sun, and Shiliang Pu.
\newblock Rpvnet: A deep and efficient range-point-voxel fusion network for lidar point cloud segmentation.
\newblock In {\em Proceedings of the IEEE/CVF International Conference on Computer Vision}, pages 16024--16033, 2021.

\bibitem{xu2023open}
Jiarui Xu, Sifei Liu, Arash Vahdat, Wonmin Byeon, Xiaolong Wang, and Shalini De~Mello.
\newblock Open-vocabulary panoptic segmentation with text-to-image diffusion models.
\newblock In {\em Proceedings of the IEEE/CVF Conference on Computer Vision and Pattern Recognition}, pages 2955--2966, 2023.

\bibitem{yang2018pixor}
Bin Yang, Wenjie Luo, and Raquel Urtasun.
\newblock Pixor: Real-time 3d object detection from point clouds.
\newblock In {\em Proceedings of the IEEE conference on Computer Vision and Pattern Recognition}, pages 7652--7660, 2018.

\bibitem{yin2021center}
Tianwei Yin, Xingyi Zhou, and Philipp Krahenbuhl.
\newblock Center-based 3d object detection and tracking.
\newblock In {\em Proceedings of the IEEE/CVF conference on computer vision and pattern recognition}, pages 11784--11793, 2021.

\bibitem{yin2021multimodal}
Tianwei Yin, Xingyi Zhou, and Philipp Kr{\"a}henb{\"u}hl.
\newblock Multimodal virtual point 3d detection.
\newblock {\em Advances in Neural Information Processing Systems}, 34:16494--16507, 2021.

\bibitem{zhou2018voxelnet}
Yin Zhou and Oncel Tuzel.
\newblock Voxelnet: End-to-end learning for point cloud based 3d object detection.
\newblock In {\em Proceedings of the IEEE conference on computer vision and pattern recognition}, pages 4490--4499, 2018.

\end{thebibliography}
\bibliographystyle{plain}

%%%%%%%%%%%%
% \include{Conference-LaTeX-template_10-17-19/exp}
% \input{exp}
%%%%%%%%%%%%

\end{document}